\documentclass[conference]{IEEEtran}
\IEEEoverridecommandlockouts
\usepackage{cite}
\usepackage{amsmath,amssymb,amsfonts}
\usepackage{algorithmic}
\usepackage{graphicx}
\usepackage{textcomp}
\usepackage{xcolor}
\usepackage{multirow}
\usepackage{color}
\usepackage{algorithm}
\def\BibTeX{{\rm B\kern-.05em{\sc i\kern-.025em b}\kern-.08em
    T\kern-.1667em\lower.7ex\hbox{E}\kern-.125emX}}
\begin{document}

\title{Towards Unsupervised Graph Completion Learning on Graphs with Features and Structure Missing \\ 
\thanks{This work was supported in part by the National Key Research and Development Program of China under Grant 2022YFF0712300, in part by the National Natural Science Foundation of China under Grant 62172177, in part by the Fundamental Research Funds for the Central Universities (HUST) under Grant 2022JYCXJJ034. $^{*}$Xinge You is the corresponding author.}}

\makeatletter
\newcommand{\linebreakand}{%
  \end{@IEEEauthorhalign}
  \hfill\mbox{}\par
  \mbox{}\hfill\begin{@IEEEauthorhalign}
}
\makeatother

% \author{\IEEEauthorblockN{Anonymous}}

\author{\IEEEauthorblockN{Sichao Fu}
\IEEEauthorblockA{\textit{School of Electronic Information and Communications} \\
\textit{Huazhong University of Science and Technology}\\
Wuhan, China \\
fusichao$\_$upc@163.com}

\and
\IEEEauthorblockN{Qinmu Peng}
\IEEEauthorblockA{\textit{School of Electronic Information and Communications} \\
\textit{Huazhong University of Science and Technology}\\
Wuhan, China \\
pengqinmu@hust.edu.cn}

\linebreakand % <------------- \and with a line-break
\IEEEauthorblockN{Yang He}
\IEEEauthorblockA{\textit{Platform Operation and Marketing Center} \\
\textit{JD Retail}\\
Beijing, China \\
landy@jd.com}

\and
\IEEEauthorblockN{Baokun Du}
\IEEEauthorblockA{\textit{Platform Operation and Marketing Center} \\
\textit{JD Retail}\\
Beijing, China \\
dubaokun@jd.com}

\linebreakand % <------------- \and with a line-break
\IEEEauthorblockN{Xinge You$^{*}$}
\IEEEauthorblockA{\textit{School of Electronic Information and Communications} \\
\textit{Huazhong University of Science and Technology}\\
Wuhan, China \\
youxg@mail.hust.edu.cn}

}

\maketitle

\begin{abstract}

In recent years, graph neural networks (GNN) have achieved significant developments in a variety of graph analytical tasks. Nevertheless, GNN’s superior performance will suffer from serious damage when the collected node features or structure relationships are partially missing owning to numerous unpredictable factors. Recently emerged graph completion learning (GCL) has received increasing attention, which aims to reconstruct the missing node features or structure relationships under the guidance of a specifically supervised task. Although these proposed GCL methods have made great success, they still exist the following problems: the reliance on labels, the bias of the reconstructed node features and structure relationships. Besides, the generalization ability of the existing GCL still faces a huge challenge when both collected node features and structure relationships are partially missing at the same time. To solve the above issues, we propose a more general GCL framework with the aid of self-supervised learning for improving the task performance of the existing GNN variants on graphs with features and structure missing, termed unsupervised GCL (UGCL). Specifically, to avoid the mismatch between missing node features and structure during the message-passing process of GNN, we separate the feature reconstruction and structure reconstruction and design its personalized model in turn. Then, a dual contrastive loss on the structure level and feature level is introduced to maximize the mutual information of node representations from feature reconstructing and structure reconstructing paths for providing more supervision signals. Finally, the reconstructed node features and structure can be applied to the downstream node classification task. Extensive experiments on eight datasets, three GNN variants and five missing rates demonstrate the effectiveness of our proposed method.

\end{abstract}


\section{Introduction}

\begin{figure}[ht]
\begin{minipage}[b]{0.495\linewidth}
  \centering
  \includegraphics[width=\linewidth]{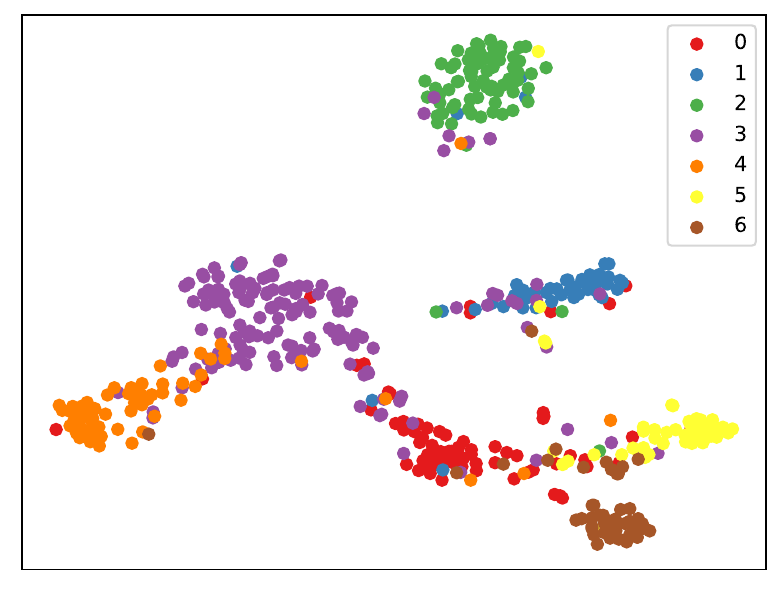}
  \centerline{(a) 15$\%$}\medskip
\end{minipage}
\begin{minipage}[b]{0.495\linewidth}
  \centering
  \includegraphics[width=\linewidth]{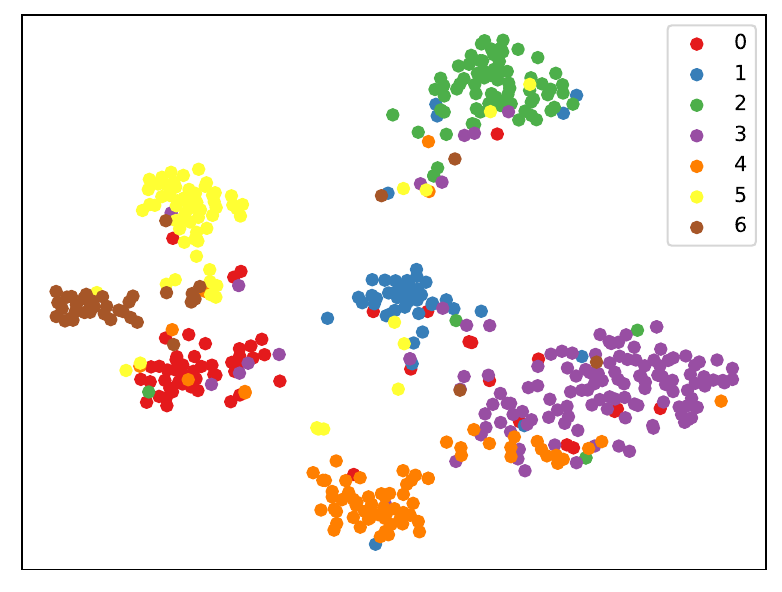}
  \centerline{(b) 35$\%$}\medskip
\end{minipage}
\begin{minipage}[b]{0.495\linewidth}
  \centering
  \includegraphics[width=\linewidth]{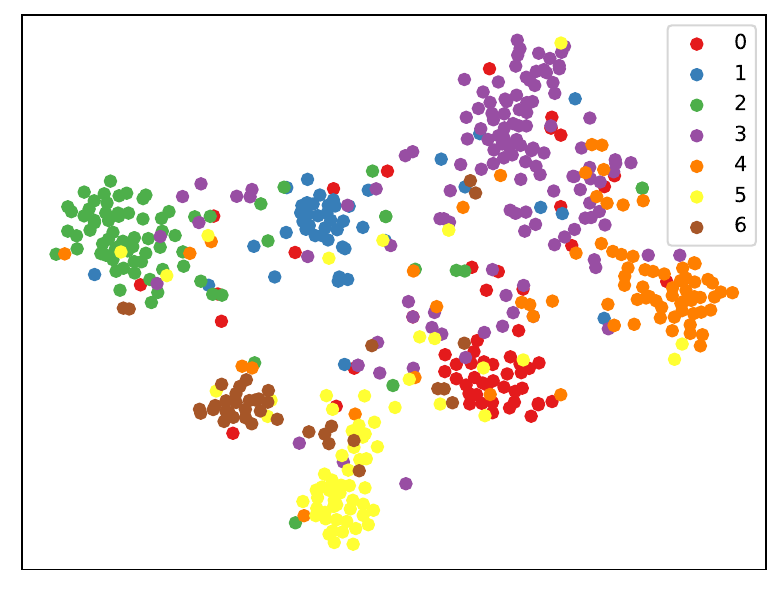}
  \centerline{(c) 55$\%$}\medskip
\end{minipage}
\begin{minipage}[b]{0.495\linewidth}
  \centering
  \includegraphics[width=\linewidth]{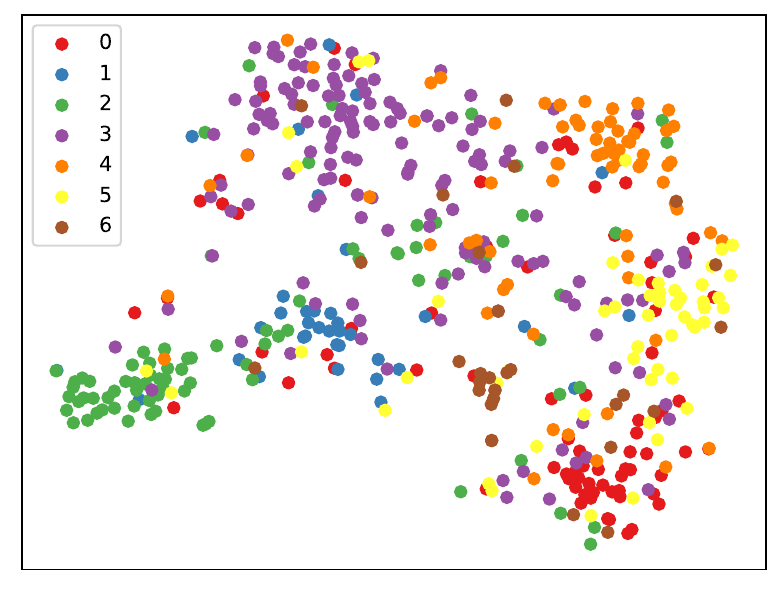}
  \centerline{(d) 75$\%$}\medskip
\end{minipage}
% \begin{minipage}[b]{0.19\linewidth}
%   \centering
%   \includegraphics[width=\linewidth]{GCN-Cora-95.pdf}
%   \centerline{(e) 95$\%$}\medskip
% \end{minipage}

\caption{t-SNE feature embeddings of classical GCN \cite{kipf2016semi} model on the Cora dataset with different rates of features and structure missing. Each color represents a different category. For t-SNE, the more obvious the discrimination of node features between different categories, the better the clustering effect. From these results, we can find that the clustering effect is getting worse and worse with the missing rates increase. It powerfully demonstrates that the generalization ability of GCN \cite{kipf2016semi} faces a huge challenge on graphs with features and structure missing, which also reveals the importance of graph completion learning.}
\label{img}
\end{figure}

With the arrival of the big data era, everything is interconnected in real life. At the same time, the awareness of digging out the potential value of people from massive data with complex interaction relationships also has become more and more powerful \cite{liu2021unified}. The emergence of the graph neural networks (GNN) \cite{pasa2022backpropagation, zhang2021affinity} provides an effective solution for the above issue. GNN has received a wide range of research interests from researchers owing to its powerful data modeling capacity in the non-Euclidean space. In recent years, a large body of GNN variants have been proposed and also achieved state-of-the-art performance in various fields, such as node classification \cite{maekawa2022beyond, ma2021learning, fu2021dynamic}, node clustering \cite{muller2023graph, miklautz2022deep, fettal2022efficient}, natural language processing \cite{yang2023semantic, yang2021bigram, yang2020targeted} and computer vision \cite{xu2022gct, fu2022adaptive, yang2019survey}.

The superior performance of the existing GNN variants highly depends on the fundamental assumption that the collected node features and structure relationships are all complete and available at any time and any place \cite{kipf2016semi, hamilton2017inductive, velivckovic2017graph, gasteiger2018predict}. However, the above assumption does not always work due to many unpredictable factors during the data collection, which seriously hurt the performance of the existing GNN on graphs with node features and structure missing. For example, (1) In a co-purchase network, businesses can obtain only the age and income privacy information of a small subset of social network users owning to data collection provision. (2) In a social network, more and more users cannot provide a full description associated with them with the improvement of safety precautions awareness. (3) Collection technique and device constraints make the collector cannot acquire complete sample features and connection relationships. Thus, the performance of the existing GNN variants will have a significant decrease when facing the above graph analytical tasks with features and structure missing. Figure 1 illustrates the t-SNE feature embeddings of classical GCN \cite{kipf2016semi} model on the Cora dataset with different rates of features and structure missing.

\begin{figure}[ht]
\begin{minipage}[b]{0.495\linewidth}
  \centering
  \includegraphics[width=\linewidth]{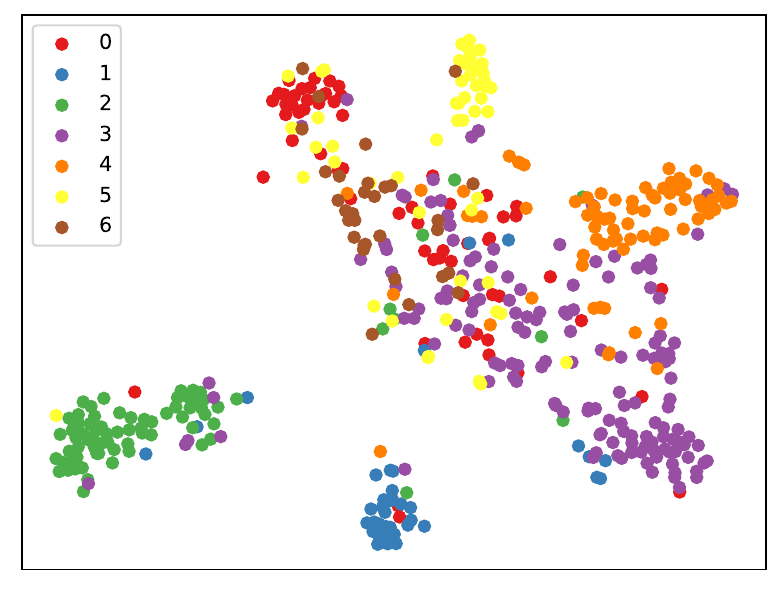}
  \centerline{(a) 10$\%$}\medskip
\end{minipage}
\begin{minipage}[b]{0.495\linewidth}
  \centering
  \includegraphics[width=\linewidth]{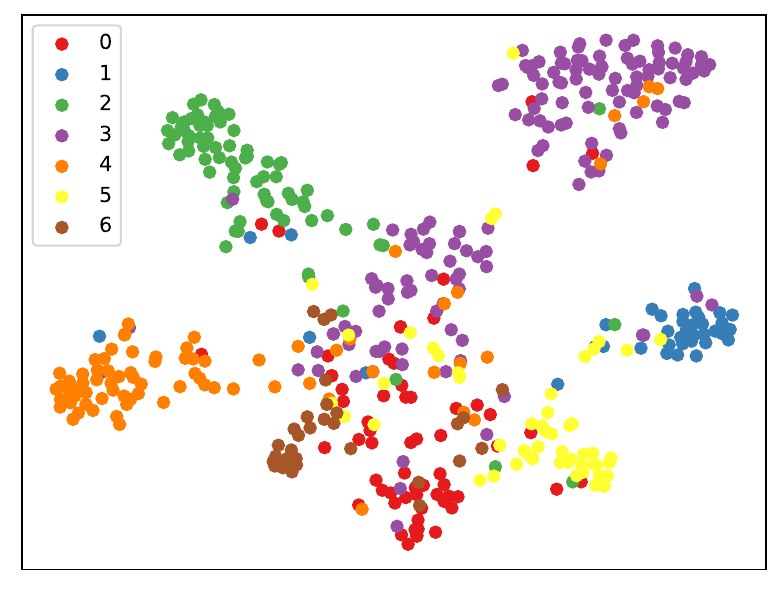}
  \centerline{(b) 20$\%$}\medskip
\end{minipage}
\begin{minipage}[b]{0.495\linewidth}
  \centering
  \includegraphics[width=\linewidth]{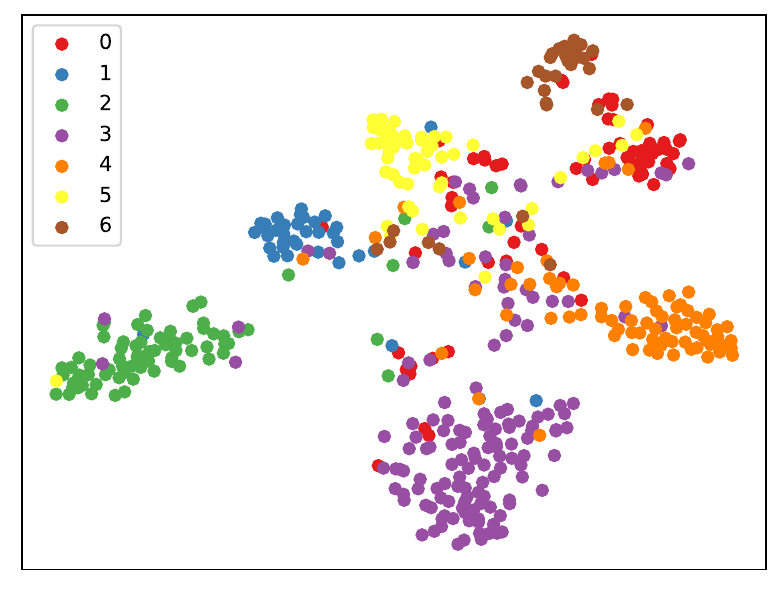}
  \centerline{(c) 30$\%$}\medskip
\end{minipage}
\begin{minipage}[b]{0.495\linewidth}
  \centering
  \includegraphics[width=\linewidth]{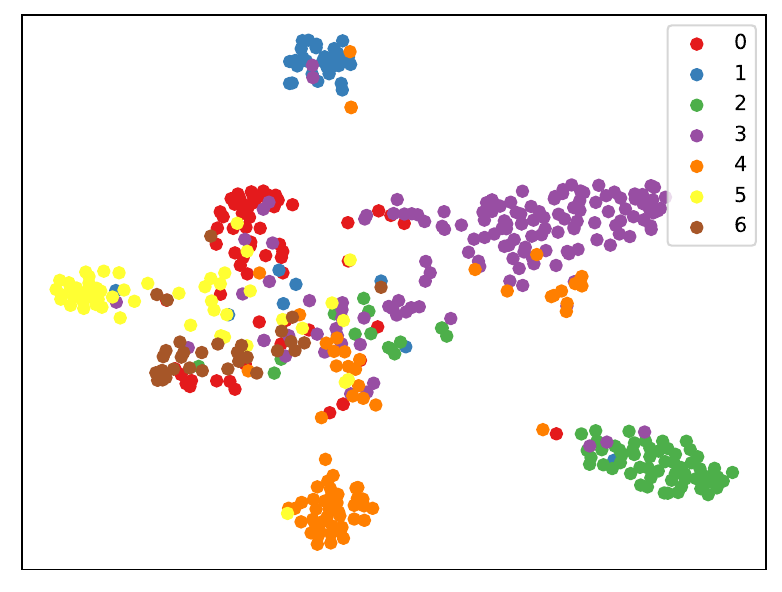}
  \centerline{(d) 40$\%$}\medskip
\end{minipage}
% \begin{minipage}[b]{0.19\linewidth}
%   \centering
%   \includegraphics[width=\linewidth]{T2-GCN-Cora-50.pdf}
%   \centerline{(e) 50$\%$}\medskip
% \end{minipage}

\caption{t-SNE feature embeddings of the recent state-of-the-art T2-GCN \cite{huo2022t2} method on the Cora dataset with different label rates of the downstream node classification task. Each color represents a different class. With the label rates increase, the better the clustering effect. This phenomenon reveals that the annotated labels from a specifically supervised task play an important role in reconstructing the missing node feature attributes and structure connection relationships.}
\label{img}
\end{figure}

To improve the generalization ability of the existing GNN variants on graphs with features and structure missing, many researchers have done massive attempts to complete the missing feature attributes or structure connection relationships for their training. These designed models are called graph completion learning (GCL) \cite{arora2020survey}. Existing GNN-based GCL models can be divided into features completion-based GCL and structure completion-based GCL methods.

Features completion-based GCL reconstructs the missing node feature attributes with the aid of a specifically supervised task according to the one-to-one matching information between the complete structure relationships and missing node features. For example, GCNMF \cite{taguchi2021graph} introduced the Gaussian mixture model to describe the node features with attribute missing, and then jointly optimized the parameters of feature reconstruction GMM and feature extraction GCN modules in an end-to-end way. FP \cite{rossi2022unreasonable} carried out the iterative diffusion between the known and unknown attributes of nodes based on minimization of the Dirichlet energy for reconstructing the missing node feature attributes. SAT \cite{chen2020learning} assumed that the node features and the corresponding structure relationships exist in the shared latent space, and then utilized the corresponding matching information for node features completion. SGHFP \cite{lei2023self} jointly used the hypergraph structure on the feature and pseudo-label space to better preserve the higher-order correlations among the known and unknown attributes. 

Structure completion-based GCL updates the wrong, uncertain and missing structure relationships along with GNN under the guidance of a specifically supervised task. For example, IDGL \cite{chen2020iterative} jointly and iteratively updated the wrong and unconnected data correlations based on the robust node embeddings learned on the GNN variants. In GAUG-M \cite{zhao2021data}, a simple graph auto-encoder algorithm including a two-layer GCN and the inner-product decoder modules is introduced to generate the high-order data correlations among nodes. DGLCN \cite{fu2021dynamic} jointly optimized the initial graph structure and the learned node embeddings at each iteration, which will stop training when the loss function satisfies the specific conditions. PTDNet \cite{luo2021learning} aimed to drop connection relationships that are unrelated to the current task by introducing the nuclear norm regularization to constrain the optimized graph structure. 

Nevertheless, these proposed GCL methods only deal with graph analytical tasks with features missing or graph analytical tasks with structure missing. The generalization ability of the existing GCL still faces a huge challenge when both collected node features and structure relationships are partially missing at the same time. To solve this problem, the recently developed T2-GCN \cite{huo2022t2} is an effective solution, which introduced the dual distillation technique to make the existing GNN variants obtain the robust node features completion and structure completion capacity at the same. Although the emerged T2-GCN \cite{huo2022t2} improves the performance of the existing GNN variants on graphs with features and structure missing to a certain degree, it still exists the following problems: (1) The reliance on labels. The existing GCL models follow a supervised learning scenario. The label numbers of a specifically supervised task are the key to deciding the performance of the existing GNN variants on the downstream tasks. Such reliance on labels from a specifically supervised task seriously limits their scope of application for the existing GCL, especially when only a few training samples and no training samples are labeled. Figure 2 illustrates the t-SNE feature embeddings of the recent state-of-the-art T2-GCN \cite{huo2022t2} method on the Cora dataset with different label rates of the downstream node classification task. (2) The bias of the reconstructed node features and structure relationships. In these supervised GCL models, the downstream node classification task only provides a small fraction of labeled samples. Thus, the nodes that provide labels and their neighbors will receive more guidance during the node features completion and structure completion, which has been demonstrated in \cite{fatemi2021slaps}. Such an imbalance phenomenon causes the bias of the reconstructed node features and structure relationships, seriously affecting the generalization ability of the existing GNN variants on the downstream task.

\begin{figure*}[ht]
    \centering
    \includegraphics[width=\linewidth]{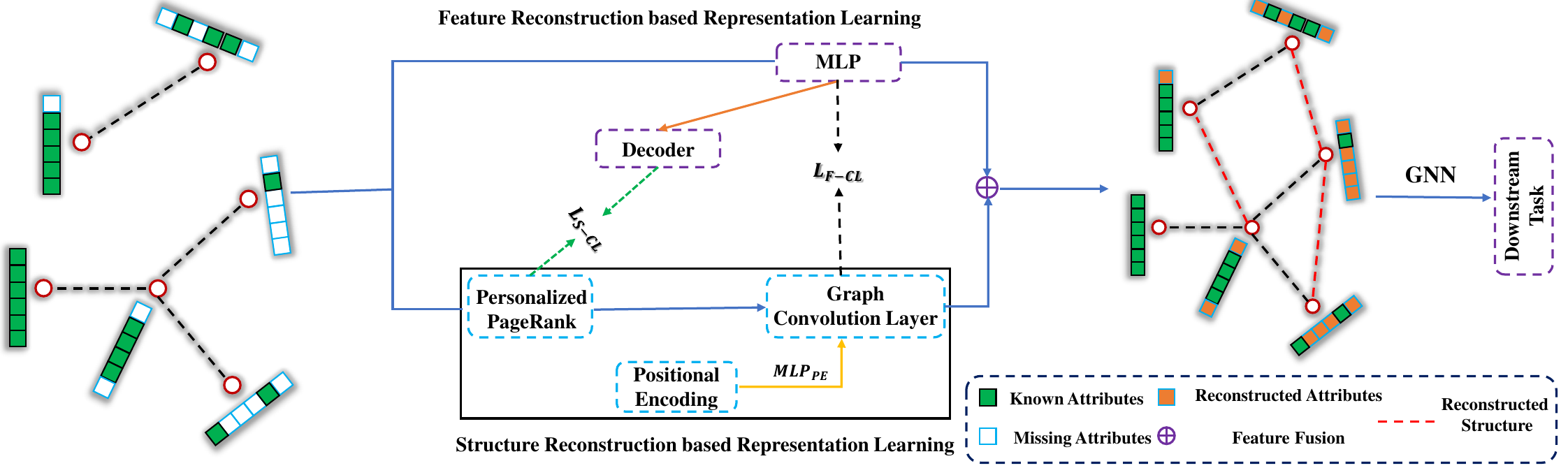}
    \caption{A diagram illustrating the proposed unsupervised graph completion learning framework on graphs with features and structure missing.}
\end{figure*}

In this paper, we propose a more effective GCL framework (UGCL) for graphs with features and structure missing under the guidance of self-supervised learning, where the missing features and structure are reconstructed by the supervision signals from the data itself. Specifically, the misleading interference between missing node features and structure caused by the message-passing scheme of GNN makes the reconstructed node features and structure relationships contain much wrong and uncertain information \cite{yang2022graph}, when both collected node features and structure relationships are partially missing at the same time. Therefore, the above mechanism-based reconstructed information will seriously hurt the perforce of the existing GNN variants on the downstream task. To avoid this issue, we separate the feature reconstruction and structure reconstruction and design its personalized model to reconstruct the missing node features or structure in the original data, respectively. For the feature reconstruction path, we introduce a deep neural network with the learnable parameters to complete the missing node feature attributes, and then a simple inner-product decoder is introduced to generate the corresponding structure relationships. For the structure reconstruction path, a simple PPNP \cite{gasteiger2018predict} model including Personalized PageRank and a two-layer GCN module is introduced to acquire the complete structure relationships and the node representations aggregating neighbor information at the same time. And then, our proposed method further introduces a dual contrastive loss on the structure level and feature level to guide node features and structure reconstruction optimization by maximizing the consistency of node representations between feature reconstructing and structure reconstructing paths as possible. As such, all nodes will acquire a fair supervision signal to avoid bias of reconstruction information and also better approximate the true data distributions of each node. After multi-iteration optimization, the reconstructed node features and structure can be combined with any existing GNN variants for the downstream node classification task.

The contributions of this work are summarized as follows:

\begin{enumerate}
\item We propose a more practical and challenging GCL paradigm, i.e. unsupervised graph completion learning on graphs with features and structure missing. To the best of our knowledge, we are the first attempt to tackle the graph analytical tasks with features and structure missing along with GNN in an unsupervised setting.

\item We propose a simple and effective unsupervised GCL framework for improving the performance of the existing GNN variants on graphs with features and structure missing. To improve model adaption, the decoupling technique and a dual contrastive loss on the structure level and feature level are introduced to guide model optimization. Our proposed UGCL also is a flexible module, which can be combined with any existing GNN variants for the downstream node classification task. 

\item Extensive experiments on eight datasets, three GNN variants and five missing rates demonstrate the effectiveness and scalability of the proposed method in comparison to the existing GCL and GNN variants.

\end{enumerate}

\section{Unsupervised Graph Completion Learning}

In this section, we elaborate that how to utilize the designed UGCL framework to enhance the generalization ability of the existing GNN variants on graphs with features and structure missing in detail. As shown in Figure 3, our proposed UGCL has a total of 4 steps. Specifically, (1) Considering the mismatch between missing node features and structure caused by the message-passing scheme of GNN, we separate the feature reconstruction and structure reconstruction, and then design two personalized methods (Feature Reconstruction / Structure Reconstruction based Representation Learning) to reconstruct its missing node feature attributes or structure relationships. For the feature reconstruction path, we introduce a two-layer multi-layer perceptron (MLP) with learnable parameters and a simple inner-product decoder module to complete the missing node features and acquire the node structure generated on the reconstructed features, respectively. (2) For the structure reconstruction path, the PPNP method including Personalized PageRank and two-layer graph convolution layer is introduced to learn the node representations, where Personalized PageRank and $MLP_{PE}$ is used to capture connection relationships between nodes and node feature attributes with spatial relations, respectively. (3) A dual contrastive loss on the structure level and feature level is designed to guide model optimization by maximizing the agreement among node representations from the different reconstructing paths. After multiple epoch optimization, we can learn its optimally reconstructed node features and structure relationships.  (4) Considering the difference between reconstructed node features from different reconstructing paths, the widely-used attention mechanism is introduced to generate robust node features. And then the reconstructed node features and structure can be applied for the optimization of GNN variants on the downstream task.

\subsection{Problem Definition}

Before describing the designed UGCL framework, we first introduce the related definitions and symbols used in the paper. $\mathbf{G} = (X, A)$ is used to describe an undirected graph, where $X = \{x_{1}, x_{2}, \cdots, x_{n} \}$ denotes the original feature attributes of all nodes and $A = \{a_{11}, a_{12}, \cdots, a_{nn} \}$ represents the connection relationships between different nodes. $x_{n}$ is the feature vector of the n-th node. In this paper, our main purpose is to reconstruct the missing node attributes and structure connection relationships from known node features and structure by our UGCL, where node labels are unavailable during the feature reconstruction and structure reconstruction process. 

\textbf{\emph{Definition \uppercase\expandafter{\romannumeral1}}}. Given an undirected graph $\mathbf{G} = (X, A)$ with the attributes missing feature matrix $X$ and connection relationships missing structure matrix $A$, the main target of the proposed UGCL is to automatically refine the original $X$ and $A$ to be the $\mathbf{G}$ with complete node features and structure, which can better describe the attributes of each node and underlying dependency relationships between different nodes. And then, the reconstructed node features and structure relationship can be applied to the training and testing of any existing GNN variants from the downstream task.

\subsection{Feature Reconstruction based Representation Learning}

The great success of most of the existing GCL methods \cite{luo2021learning, taguchi2021graph} highly rely on the assumption that either the original node feature attributes are missing or the structure relationships are missing. With the above assumption, the existing GCL utilizes the one-to-one matching information between node features and structure to reconstruct the missing node feature attributes or structure connection information. Nevertheless, this assumption does not always work when both collected node features or structure relationships are partially missing at the same time. The main reason for the above issue is the node feature attributes missing and structure connection information missing occur at the same time. In this case, the message-passing scheme of GNN variants is negative and even degradation their task performance owning to a misleading match between missing node features and structure, which has been demonstrated in \cite{fatemi2021slaps}.

In this part, we introduce a two-layer parameterized multi-layer perceptron (MLP) to reconstruct the missing node feature attributes by the powerful learning ability of the deep learning model, i.e.

\begin{equation}
x_n^{FR} = 
\begin{cases}
f_{\theta}(x_n) & \text{$x_n$ exists missing attribute $x_{nd}$} \\
x_n & \text{otherwise}
\end{cases}
\end{equation}

where $x_n$ denotes the original node feature vector with feature attributes missing. $X_{nd}$ is the d-th column missing node attribute of the n-th node. $f_{\theta}()$ denotes the a two-layer MLP with the learnable parameter $\theta$. $x_n^{FR}$ represents the complete node features. With the node features matrix $x_n^{FR}$ with complete attributes, we introduce a simple inner-product decoder module to capture underlying dependency between nodes, which can be written as follows:

\begin{equation}
A^{FR} =  \sigma(X^{FR}{X^{FR}}^{T}) 
\end{equation}

where $T$ is the transposition operation of the matrix and $A^{FR}$ denotes the generated structure information in the feature reconstruction path. $\sigma$ is the activation function.

\subsection{Structure Reconstruction based Representation Learning}

For the existing GNN variants, the key to their superior performance is that the provided graph structure whether can describe local geometry distributions between nodes properly. Therefore, our main target is how to better reconstruct the missing structure from known structure relationships. To solve this issue, we introduce a two-layer PPNP \cite{gasteiger2018predict} model including Personalized PageRank and the graph convolution layer modules to reconstruct the missing structure and learn the corresponding node representations, i.e. 

\begin{equation}
Z^{SR} = \hat{A}^{SR}(\sigma (\hat{A}^{SR}  \hat{X}_{PE} W^{(0)})) W^{(1)}
\end{equation}

where $Z^{SR}$ denotes the learned node representations in the structure reconstruction path by a two-layer PPNP. $\hat{A}^{SR}$ is the reconstructed structure connection relationships in the structure reconstruction path. $\hat{X}_{PE}$ denotes the positional encoding vector of each node. $W^{(0)}$ and $W^{(1)}$ represent the weight parameters matrix of the first layer and the second layer PPNP, respectively. For $\hat{A}^{SR}$, it is generated by the Personalized PageRank with kNN sparsification. Personalized PageRank \cite{haveliwala2002topic} utilized the widely-used restarted random walk to generate the ranking scores between each node and its neighbors according to the original graph structure $A$. Starting from the known structure relationships of $A$, Personalized PageRank iteratively updated the missing structure relationships of $A$ using the following formulation:

\begin{equation}
 A_{t+1}^{SR} = (1 - \alpha)(A + I_n)A_{t}^{SR} + \alpha I_n
\end{equation}

where $A_{t}^{SR}$ denotes the graph structure $A$ at the timestamp $t$ and $I_n$ denotes the identity matrix. $\alpha$ is the reset probability. After multi-time iterations, the closed-form solution of the above Personalized PageRank can be written as follows:

\begin{equation}
 A^{SR} = \alpha (I_n - (1 - \alpha)(A + I_n))^{-1}
\end{equation}

To improve the computing efficiency of PPNP, the well-known K-NearestNeighbor sparsification technique is introduced to remove some unimportant scores in $ A^{SR}$, i.e.

\begin{equation}
\hat{A}_{ij}^{SR} = 
\begin{cases}
A_{ij}^{SR} & \text{ $j \in kNN(i)$ } \\
0 & \text{ $j \notin kNN(i)$ }
\end{cases}
\end{equation}

where $kNN(j)$ denotes the set of top-k neighbors of the i-th node. Simple depending on the one-to-one matching information between the complete structure relationships and missing node features still cannot guarantee the learned node representation is the most representative. To further enhance the effectiveness of the learned node representation $Z^{SR}$, we introduce the positional encoding \cite{shaw2018self} method to encode the relative spatial relationships of each node. And then, a simple $MLP_{PE}$ $f_{\vartheta}()$ is introduced to make linear transformation on the one-hot positional encoding vector of each node as the initial node feature attributes, i.e. 

\begin{equation}
 \hat{X}_{PE} = f_{\vartheta}(X_{PE})
\end{equation}

\subsection{Model Training and Downstream Tasks Testing}

To reconstruct the effective node features and structure relationships for the downstream task, we design a dual contrastive loss on the structure level and feature level to maximize the mutual information of node representations from feature reconstruction and structure reconstruction paths for guiding model optimization. The overall objective function $\mathcal{L}$ can be written as follows:

\begin{algorithm}[tb]
\caption{Proposed UGCL for Node Classification on Graphs with Features and Structure Missing}
\label{alg:algorithm}
\textbf{Input}: node features matrix $X$, adjacency matrix $A$, adopted GNN variant in the downstream task. \\
\textbf{Parameters}: hidden units, output units, learning rate, weight decay, dropout rate, reset probability $\alpha$ and the number of neighbors $k$, number of epochs $E$. \\
\textbf{Output}: reconstructed node features matrix $\hat{X}$, reconstructed adjacency matrix $\hat{A}^{SR}$ and the test performance of the the downstream task.
\begin{algorithmic}[1] %[1] enables line numbers
\STATE Initialize parameters $\theta$, $\vartheta$, $W^{(0)}$ and $W^{(1)}$. 
\FOR{$ e=0 \rightarrow E-1 $}
\STATE Calculate $X^{FR}$ by Eq. (1) and $A^{FR}$ by Eq. (2);
\STATE Calculate $\hat{A}^{SR}$ by Eq. (4) - Eq. (6) and $\hat{X}_{PE}$ by Eq. (7); After then, generate the corresponding node representations $Z^{SR}$ by Eq. (3);
\STATE Calculate the dual contrastive loss $\mathcal{L}$ by Eq. (8);
\STATE Update parameters $\theta$, $\vartheta$, $W^{(0)}$ and $W^{(1)}$ by utilizing gradient descent;
\ENDFOR \\
\STATE Obtain the optimally reconstructed node features matrix $\hat{X}$ by Eq. (9) - Eq. (10) and adjacency matrix $\hat{A}^{SR}$;
\STATE Update $\varphi$, $\varphi^{'}$, $W_{F^2}$, $W_{F^2}^{'}$ and parameters of adopted GNN variant in the downstream task;
\STATE \textbf{Return} $\hat{X}$, $\hat{A}^{SR}$ and the test performance of the downstream task.
\end{algorithmic}
\end{algorithm}

\begin{small}
\begin{equation}
\begin{split}
\mathcal{L} & =  \min \limits_{\theta, \vartheta, W^{(0)}, W^{(1)}} (\mathcal{L}_{F-CL} + \mathcal{L}_{S-CL}) \\
& =  \min \limits_{\theta, \vartheta, W^{(0)}, W^{(1)}} \sum_{i=1}^{n} \log \frac{e^{cos(X_i^{FR}, Z_i^{SR})/t}}{e^{cos(X_i^{FR}, Z_i^{SR})/t} + \sum_{i \neq j} e^{cos(X_i^{FR}, Z_j^{SR})/t}}   \\
& + \min \limits_{\theta, \vartheta, W^{(0)}, W^{(1)}}  \sum_{i=1}^{n} \log \frac{e^{cos(A_i^{FR}, \hat{A}_i^{SR})/t}}{e^{cos(A_i^{FR}, \hat{A}_i^{SR})/t} + \sum_{i \neq j} e^{A_i^{FR}, \hat{A}_i^{SR})/t}} \\
\end{split}
\end{equation}
\end{small}

where $\mathcal{L}_{F-CL}$ denotes the contrastive loss on the feature level, which aims to maximize the intra-class mutual information among the same node from the different reconstruction paths and minimize the inter-class consistency of different nodes from the different reconstruction paths. $\mathcal{L}_{S-CL}$ is the contrastive loss on the structure level, which pulls closer the structure information of the same node and pushes different nodes away. $cos(,)$ denotes the cosine similarity among nodes and $t$ is a scaler temperature. By minimizing $\mathcal{L}$ during the UGCL training process, we can obtain the optimal node features matrix and adjacency matrix. Considering the difference between the reconstructed node features $X^{FR}$ and the learned node representations $Z^{SR}$ aggregating the reconstructed structure information $\hat{A}^{SR}$, the widely-used attention mechanism is introduced to analyze the importance of $Z^{SR}$ and $\hat{A}^{SR}$ for the current downstream task, i.e.

\begin{equation}
\begin{cases}
{\gamma}_0 = tanh(f_\varphi^{T}(X^{FR}) W_{F^2} X^{FR}) &  \\
{\gamma}_1 = tanh(f_{\varphi^{'}}^{T}(Z^{SR}) W_{F^2}^{'} Z^{SR}) & 
\end{cases}
\end{equation}

\begin{equation}
\hat{X} =  \frac{exp({\gamma}_0) }{exp({\gamma}_0) + exp({\gamma}_1)} X^{FR} + \frac{exp({\gamma}_1) }{exp({\gamma}_0) + exp({\gamma}_1)} Z^{SR}
\end{equation}

where $f_\varphi^{T}(X^{FR})$ and $f_{\varphi^{'}}^{T}(Z^{SR})$ denote the transposition of the feature embeddings after the linear transformation of MLP $f_\varphi()$ or $f_\varphi^{'}()$, respectively. $\varphi$, $\varphi^{'}$, $W_{F^2}$, $W_{F^2}^{'}$ are the trainable parameter of the above feature fusion module. $\frac{exp({\gamma}_0) }{exp({\gamma}_0) + exp({\gamma}_1)}$ and $\frac{exp({\gamma}_1) }{exp({\gamma}_0) + exp({\gamma}_1)}$ denote the importance of $Z^{SR}$ and $\hat{A}^{SR}$ for the current downstream task by the normalization operation. Finally, we can obtain the optimally reconstructed node features matrix $\hat{X}$ by their weight combination. During the downstream task testing, the optimally reconstructed node features matrix $\hat{X}$ and adjacency matrix $\hat{A}^{SR}$ can be combined with any existing GNN variants for GNN variants' optimization, which can effective improve the generalization ability of the existing GNN variants on graphs with features and structure missing. The overall algorithm flowchart of the proposed UGCL method is shown in Algorithm 1.

\section{Experiment}

\subsection{Datasets}

To demonstrate the effectiveness and scalability of the proposed UGCL framework, we conduct extensive experiments on eight real-world graph benchmarks, including Citation network \cite{yang2016revisiting} (Cora, Citeseer, and Pubmed), Webpage network \cite{pei2020geom} (Cornell, Texas, and Wisconsin) and Wikipedia network \cite{rozemberczki2021multi} (Chameleon and Squirrel).

For the Citation network, each node denotes some kind of paper or document, which is described by many 0/1 bag-of-words vectors. Each edge shows the citation relationships among nodes. Specifically, Cora is composed of 2708 scientific publications with 7 categories, including case-based, probabilistic methods, genetic algorithms, neural networks, rule learning, theory, and reinforcement learning. The entire database contains 5278 different types of citation links and collects 1433-dimensional feature vectors for describing the corresponding node attributes. Citeseer contains a total of 3327 documents with a size of 3703 collected from 6 classes, such as information retrieval, artificial intelligence, database, agents, human-computer interaction, and machine language. The edges are described by 4676 citation links between 3327 papers. Pubmed consists of a total of 19717 scientific publications, which are categorized into 3 categories with a size of 500, such as diabetes mellitus type 1, diabetes mellitus experimental, and diabetes mellitus type 2. It is also composed of 44327 citation links.

Cornell, Texas, and Wisconsin datasets are collected from the web pages of the corresponding universities by Carnegie Mellon University. Each node and edge denote the different types of web pages and the hyperlinks associated with them. The web page of all Webpage network is described by many 0/1 bag-of-words vectors with the size of 1703, which are divided into 5 classes including student, project, course, staff, and faculty. To be specific, Cornell and Texas contain 183 web pages. Wisconsin is composed of a total of 251 web pages.

\begin{table}[t]
	\caption{Dataset Statistics.}
	\centering
    \begin{tabular}{c | c | c | c | c | c }
    \hline
    Dataset & Data Type & Nodes & Edges & Classes & Features \\
    \hline
    Cora  & Citation Network & 2708 & 5278 & 7 & 1433  \\
    Citeseer  & Citation Network & 3327 & 4676 & 6 & 3703\\
    Pubmed & Citation Network & 19717 & 44327 & 3 & 500  \\

    Cornell  & Webpage Network & 183 & 295 & 5 & 1703   \\
    Texas  & Webpage Network & 183 & 295 & 5 & 1703   \\
    Wisconsin & Webpage Network & 251 & 499 & 5 & 1703   \\

    Chameleon  & Wikipedia Network & 2277 & 31421 & 5 & 2325  \\
    Squirrel  & Wikipedia Network & 5201 & 198493 & 5 & 2089  \\
    \hline
    
    \end{tabular}
	\label{datasets1}	
\end{table}

Different from the Webpage network, each node from Chameleon and Squirrel is collected from the Wikipedia web pages. According to the average monthly traffic of each web page, all web pages are further divided into 5 classes. For example, Chameleon is composed of 2277 web pages and 31421 hyperlinks associated with each node. Each node contains 2325 0/1 bag-of-words feature vectors. Squirrel consists of a total of 198493 citation links, which is described by 5201 web pages with a size of 2089. The brief descriptions of the above-mentioned datasets are summarized in Table \uppercase\expandafter{\romannumeral1}.

\subsection{Experimental settings}

In our experiments, 60 $\%$ nodes per class are randomly chosen for the training set, 20 $\%$ nodes per class as the validation set, and the rest for the testing set. To be a fair comparison, both the comparison methods and our proposed UGCL all randomly mask 30 $\%$ node feature attributes and structure connection relationships for model evaluation. To avoid any bias introduced by the random partitioning of data, we carry out all models ten times independently to report their average accuracy. In the feature reconstructing and structure reconstructing paths, we adopt a two-layers Multi-layer Perceptron (MLP) and a two-layer PPNP to reconstruct the missing node features and structure in turns. The dimension of hidden units and output layer is searched in $\{64, 128, 256, 512, 1024\}$. We search the initial learning rate and weight decay from $0.1$ to $0.00009$. The reset probability $\alpha$ and the number of neighbors $k$ in PPNP are tuned from $0.1$ to $0.9$ and amongst $\{0, 5, 10, 15, 20, 30, 40, 50\}$, respectively. The dropout rate is chosen from 0.1 to 0.9. In downstream semi-supervised node classification tasks, we follow the same labeled node sampling strategy as described in \cite{huo2022t2} and the layer numbers of the adopted GNN variants are set to 2. All GNN variants use their optimal parameter to validate the effectiveness of the existing GCL and our UGCL methods.

\begin{table*}[tb]
    \caption{Experiment comparison with the state-of-the-art GCL methods on node classification task with 30$\%$ missing rate. For a fair comparison, all supervised GCL regard GCN as the backbone of the downstream task. The best and second-best results are highlighted in purple-underline and blue-underline.}
    \centering
    \resizebox{\linewidth}{!}{
    \begin{tabular}{c | c | c | c c c c c c c c c } 
    
    \hline
    &  Available Data for GCL  & Methods  & Cora & Citeseer & Pubmed & Cornell & Texas & Wisconsin & Chameleon & Squirrel \\

    \hline
     \multirow{4}*{-----} & \multirow{4}*{-----} & GCN (ICLR 2017)  \cite{kipf2016semi} & 82.77 &68.84 & 83.51 & 48.90 & 46.48 & 50.39 & 58.37 & 40.09 \\

    & & GraphSAGE (NeurIPS 2017)  \cite{hamilton2017inductive} &82.93 & 70.96 &  \textcolor{blue}{\underline{84.96}} & 64.59 &  \textcolor{purple}{\underline{70.81}} & 69.41 & 59.05 & 41.86 \\

    & & GAT (ICLR 2018)  \cite{velivckovic2017graph} &83.54 & 70.00 & 84.33 & 55.67 & 50.27 & 53.72 & 57.56 & 36.58 \\

    & & APPNP (ICLR 2019)  \cite{gasteiger2018predict} &83.48 & 70.61 & 78.77 & 55.94 & 47.83 & 53.52 & 44.18 & 28.57 \\ \cline{1-11}

    \multirow{6}*{Supervised} & \multirow{7}*{X, A, Y} & IDGL (NeurIPS 2020)  \cite{chen2020iterative} &79.88 & 65.80 & 82.60 & 54.05 & 62.16 & 60.78 & 48.68 & 33.11 \\

    & & PTDNet (WSDM 2021)  \cite{luo2021learning} &74.93 & 72.29 & 82.49 & 64.87 & 61.62 & 73.14 & 48.42 & 30.45 \\

    & & GCNMF (FGCS 2021)  \cite{taguchi2021graph} &83.08 & 71.75 & 67.08 & 55.43 & 57.28 & 53.02 & 47.75 & 30.56 \\

    & & singleT (AAAI 2023)  \cite{huo2022t2} &80.19 & 67.22 & 81.03 & 54.76 & 53.81 & 57.10 & 55.93 & 40.32 \\

    & & T2-GCN-online (AAAI 2023)  \cite{huo2022t2} &80.11 & 65.31 & 83.04 & 74.01 & 60.53 & 69.42 & 54.22 &  \textcolor{purple}{\underline{45.19}} \\

    & & T2-GCN (AAAI 2023)  \cite{huo2022t2} & \textcolor{blue}{\underline{83.84}} & \textcolor{blue}{\underline{72.32}} & 84.85 & 67.29 & 65.40 & \textcolor{blue}{\underline{73.39}} & \textcolor{purple}{\underline{60.81}} &  \textcolor{blue}{\underline{44.33}} \\

    \hline
    \multirow{2}*{Unsupervised} & \multirow{2}*{X, A} & SAT (TPAMI 2022)  \cite{chen2020learning} & 64.59 & 45.87 & - & \textcolor{blue}{\underline{76.68}} & 63.89 & 67.49 & 56.57 & 35.23 \\
    
     &  & UGCL-GCN   &  \textcolor{purple}{\underline{85.62}} &  \textcolor{purple}{\underline{74.15}} & \textcolor{purple}{\underline{85.74}} &  \textcolor{purple}{\underline{79.51}} & \textcolor{blue}{\underline{70.69}} &  \textcolor{purple}{\underline{75.85}} & \textcolor{blue}{\underline{60.76}} & 42.84 \\

    \hline
    \end{tabular}}
    \label{table:cifar}
\end{table*}

\begin{table*}[ht]
    \caption{Ablation experiments on node classification task with 30$\%$ missing rate. The best results are highlighted in purple-underline.}
    \centering
    \resizebox{\linewidth}{!}{
    \begin{tabular}{c | c c c c c c c c c } 
    \hline
    Methods  & Cora & Citeseer & Pubmed & Cornell & Texas & Wisconsin & Chameleon & Squirrel \\

    \hline

    UGCL-GCN (w/o $X^{FR}$) & 71.29 & 72.40 & 83.70 & 62.07 & 58.62 & 53.66 & 30.52 & 33.21 \\
    
    UGCL-GCN (w/o $\hat{A}^{SR}$ and $Z^{SR}$ ) & 84.65 & 68.35 & 84.85 & 68.28 & 64.14 & 74.14 & 57.06 & 40.93 \\

    UGCL-GCN (w/o Eq. (9) - Eq. (10)) & 84.63 & 71.56 & 85.25 & 62.76  & 65.86 & 74.15 & 55.72 & 40.16 \\

    UGCL-GCN (w/o $\mathcal{L}_{S-CL}$) & 84.08 & 71.16 & 85.11 & 67.59 & 61.73 & 73.90 & 57.17 & 38.15 \\

    UGCL-GCN   &  \textcolor{purple}{\underline{85.62}} &  \textcolor{purple}{\underline{74.15}} &  \textcolor{purple}{\underline{85.74}} &  \textcolor{purple}{\underline{79.51}} &  \textcolor{purple}{\underline{70.69}} &  \textcolor{purple}{\underline{75.85}} &  \textcolor{purple}{\underline{60.76}} &  \textcolor{purple}{\underline{42.84}} \\

    \hline
    \end{tabular}
    }
    \label{table:cifar}
\end{table*}

\subsection{Comparison with State-of-the-art GCL}

To demonstrate the effectiveness of the proposed UGCL in tackling semi-supervised node classification with features and structure missing, we report the average accuracy of the four representative GNN variants (GCN, GraphSAGE, GAT, and APPNP), six supervised GCL methods (IDGL, PTDNet, GCNMF, singleT, T2-GCN (online) and T2-GCN), one unsupervised GCL model (SAT) and our proposed UGCL in Table \uppercase\expandafter{\romannumeral2}. For a fair comparison, all compared GCL methods adopt the same GNN variant (GCN) during the downstream task evaluation process.
\begin{enumerate}
\item From these results, we can find that the supervised GCL methods outperform the above GNN variants in most cases, especially for the GCN model. For example, the recent T2-GNN improves average accuracy by 1.07 $\%$, 3.48$\%$, 1.34$\%$, 18.39$\%$, 18.92$\%$, 23$\%$, 2.44$\%$ and 4.24$\%$ in comparison to the same GNN variant (GCN), respectively. These results further demonstrate that the reconstructed node features and structure connection relationships by GCL can enhance the generalization ability of the existing GNN variants on graphs with feature and structure missing to some extent.

 \item Secondly, when generalizing the above features completion-based GCL (GCNMF and SAT) and structure completion-based GCL (IDGL and PTDNet) methods to deal with semi-supervised node classification task with features and structure missing at the same time, the performance of these proposed methods has a huge drop compared to the experiment results of the original paper and even lower than the adopted GNN variant. This phenomenon shows that the misleading interference between missing node features and structure caused by the message-passing scheme of GNN may have a negative influence when both collected node features or structure relationships are partially missing at the same time. For example, GCN outperforms the IDGL and GCNMF on 3 out of 8 benchmarks (Pubmed, Chameleon and Squirrel) by 0.91$\%$, 9.69$\%$ and 6.98$\%$, 16.43$\%$, 10.62$\%$ and 9.53$\%$ improvement, respectively. 

\item Compared with the above supervised GCL and GNN variants, our proposed UGCL achieves a novel state-of-the-art performance on 5 out of 8 benchmarks without the aid of labels from the downstream node classification task. For example, our UGCL obtain huge gains of 1.78 $\%$, 1.83$\%$, 0.78$\%$, 2.83$\%$, and 2.46$\%$ in comparison to the corresponding best performance on each dataset, such as T2-GCN (Cora and Citeseer), GraphSAGE (Pubmed), SAT (Cornell) and T2-GCN (Wisconsin). These results, on the one hand, validate the effectiveness of the proposed UGCL, on the other hand, reveal that reasonably utilizing the supervision signal from the data itself also guarantees the accuracy of the reconstructed node features and structure connection relationships, apart from the label information.

\end{enumerate}

\begin{table*}[tb]
    \caption{Experiment comparison with the state-of-the-art GCL methods on node classification task with with different rates of features and structure missing. The best results are highlighted in purple-underline.}
    \centering
    \resizebox{\linewidth}{!}{
    \begin{tabular}{c | c | c c c c c c c c c } 
    \hline
    Missing Rates  & Methods  & Cora & Citeseer & Pubmed & Cornell & Texas & Wisconsin & Chameleon & Squirrel \\

    \hline
   \multirow{3}*{15$\%$} & GCN  & 83.64 & 68.82 & 86.06 & 48.11 & 50.27 & 51.57 & 62.50 & 41.76 \\

    &  T2-GCN & 85.51 & 72.17 & 86.20 & 73.51 & 70.54 & 76.27 & 63.10 & 40.86 \\

     & UGCL-GCN   & \textcolor{purple}{\underline{87.20}} & \textcolor{purple}{\underline{75.17}} & \textcolor{purple}{\underline{87.38}} & \textcolor{purple}{\underline{76.21}} & \textcolor{purple}{\underline{73.10}} & \textcolor{purple}{\underline{80.25}} & \textcolor{purple}{\underline{63.27}} & \textcolor{purple}{\underline{44.20}} \\

     \hline
     \multirow{3}*{35$\%$} & GCN  & 80.61 &67.01 & 84.27 & 47.84 & 48.65 & 51.17 & 58.33 & 40.06 \\

     &  T2-GCN & 82.61 & 68.75 & 84.33 & 69.73 & 68.92 & 71.18 & 58.55 & 39.55 \\

     & UGCL-GCN   & \textcolor{purple}{\underline{83.98}} & \textcolor{purple}{\underline{72.23}} & \textcolor{purple}{\underline{84.86}} & \textcolor{purple}{\underline{70.69}} & \textcolor{purple}{\underline{72.75}} & \textcolor{purple}{\underline{75.37}} & \textcolor{purple}{\underline{60.00}} & \textcolor{purple}{\underline{41.40}} \\

      \hline
     \multirow{3}*{55$\%$} & GCN  & 75.17 &61.77 & 78.01 & 47.03 & 47.30 & 50.59 & 54.06 & 36.43 \\

     &  T2-GCN & 76.60 & 63.02 & 79.31 & 64.06 & 60.81 & 67.06 & 54.17 & 36.93 \\

     & UGCL-GCN   & \textcolor{purple}{\underline{77.80}} & \textcolor{purple}{\underline{65.71}} & \textcolor{purple}{\underline{80.48}} & \textcolor{purple}{\underline{67.24}} & \textcolor{purple}{\underline{62.42}} & \textcolor{purple}{\underline{72.43}} & \textcolor{purple}{\underline{55.67}} & \textcolor{purple}{\underline{37.01}} \\

      \hline
      \multirow{3}*{75$\%$} & GCN  & 61.17 & 50.45 & 68.53 & 54.06 & 46.76 & 50.39 & 45.74 & 31.97 \\

      &  T2-GCN & 63.08 & 51.53 & 72.98 & 62.70 & 57.30 & 60.20 & 46.71 & 32.31 \\

      & UGCL-GCN   & \textcolor{purple}{\underline{63.49}} & \textcolor{purple}{\underline{51.97}} & \textcolor{purple}{\underline{74.04}} & \textcolor{purple}{\underline{65.52}}  & \textcolor{purple}{\underline{61.73}} & \textcolor{purple}{\underline{70.73}} & \textcolor{purple}{\underline{48.66}} & \textcolor{purple}{\underline{34.26}} \\

      \hline
      \multirow{3}*{95$\%$} & GCN  & 32.58 &25.66 & 47.16 & 44.05 & 39.46 & 40.20 & 28.51 & 24.58 \\

      &  T2-GCN & 33.28 & 26.41 & 48.55 & 47.03 & 46.22 & 43.14 & 29.85 & 26.58 \\

      & UGCL-GCN   & \textcolor{purple}{\underline{33.74}} & \textcolor{purple}{\underline{27.28}} & \textcolor{purple}{\underline{49.14}} & \textcolor{purple}{\underline{62.76}} & \textcolor{purple}{\underline{48.28}} & \textcolor{purple}{\underline{54.39}} & \textcolor{purple}{\underline{32.62}} & \textcolor{purple}{\underline{27.02}} \\
     
      \hline
    
    \end{tabular}
    }
    \label{table:cifar}
\end{table*}

\begin{table*}[ht]
    \caption{Experiment comparison with the state-of-the-art GCL methods on node classification task with 30$\%$ missing rate and the different backbone of the downstream task. The best and second-best results are highlighted in purple-underline and blue-underline.}
    \centering
    \resizebox{\linewidth}{!}{
    \begin{tabular}{c | c | c | c c c c c c c c c } 
    \hline
    &  Available Data for GCL  & Encoders  & Cora & Citeseer & Pubmed & Cornell & Texas & Wisconsin & Chameleon & Squirrel \\

    \hline
    \multirow{4}*{-----} & \multirow{4}*{-----} & GCN  & 82.77 &68.84 & 83.51 & 48.90 & 46.48 & 50.39 & 58.37 & 40.09 \\

    & & GraphSAGE  &82.93 & 70.96 & 84.96 & 64.59 & 70.81 & 69.41 & 59.05 & 41.86 \\

    & & GAT &83.54 & 70.00 & 84.33 & 55.67 & 50.27 & 53.72 & 57.56 & 36.58 \\ \cline{1-11}

    % & & APPNP &83.48 & 70.61 & 78.77 & 55.94 & 47.83 & 53.52 & 44.18 & 28.57 \\ \cline{1-11}

    \multirow{4}*{Supervised} & \multirow{4}*{X, A, Y} &  T2-GCN &83.84 & 72.32 & 84.85 & 67.29 & 65.40 & 73.39 & 60.81 & 44.33 \\

    & & T2-GraphSAGE & \textcolor{blue}{\underline{84.62}} & 71.39 & 85.56 & 75.40 & \textcolor{purple}{\underline{78.10}} & 82.54 & \textcolor{purple}{\underline{62.89}} & \textcolor{blue}{\underline{45.78}} \\

     & & T2-GAT & 84.46 & 72.42 & 85.07 & 64.86 & 61.35 & 62.35 & 58.39 & 38.87 \\

     % & & T2-APPNP &84.08 & 71.77 & 83.36 & 66.48 & 57.83 & 66.27 & 50.41 & 36.93 \\
     \hline

    \multirow{4}*{Unsupervised} & \multirow{4}*{X, A} & UGCL-GCN  & \textcolor{purple}{\underline{85.62}} & \textcolor{blue}{\underline{74.15}} & \textcolor{blue}{\underline{85.74}} & 79.51 & 70.69 & 75.85 & 60.76 & 42.84 \\

    &  & UGCL-GraphSAGE & 84.05 & \textcolor{purple}{\underline{74.76}} & \textcolor{purple}{\underline{85.81}} & \textcolor{purple}{\underline{83.45}} & \textcolor{blue}{\underline{76.90}} & \textcolor{blue}{\underline{89.02}} & \textcolor{blue}{\underline{62.74}} & 45.32 \\

     &  & UGCL-GAT   & 83.94 & 72.96 & 84.80 & \textcolor{blue}{\underline{80.69}} & 71.72 & \textcolor{purple}{\underline{89.26}} & 62.71 & \textcolor{purple}{\underline{48.97}} \\

     % &  & UGCL-APPNP  & 0 & 0 & 0 & 0 & 0 & 0 & 0 & 0 \\

    \hline
    \end{tabular}
    }
    \label{table:cifar}
\end{table*}

\subsection{Ablation experiments}

In Table \uppercase\expandafter{\romannumeral3}, we conduct extensive experiments on eight benchmarks to analyze the influence of each component. UGCL-GCN (w/o $X^{FR}$) and UGCL-GCN (w/o $\hat{A}^{SR}$ and $Z^{SR}$) denote without utilizing the reconstructed node feature attributes (node features and structure) from the feature (structure) construction path to generate the optimal node features for downstream node classification. UGCL-GCN (w/o Eq. (9) - Eq. (10)) utilizes only the simple concatenation operation to fuse the reconstructed node features from different construction paths for generating the optimal node features. From the experimental results of the first, second and third columns, UGCL-GCN (w/o Eq. (9) - Eq. (10)) outperforms UGCL-GCN (w/o $X^{FR}$) and UGCL-GCN (w/o $\hat{A}^{SR}$ and $Z^{SR}$) in most cases, as the former fully considers the diversified difference of the reconstructed node feature attributes from the feature construction and structure construction paths. Although UGCL-GCN (w/o Eq. (9) - Eq. (10)) improves the classification performance of the existing GNN variants on graphs with features and structure missing, it is difficult to guarantee the generated node feature attributes and structure connection relationships are optimal for the current node classification task. Compared with UGCL-GCN (w/o Eq. (9) - Eq. (10)), our proposed UGCL-GCN achieves 0.99 $\%$, 2.59$\%$, 0.49$\%$, 16.75$\%$, 4.83$\%$, 1.7$\%$, 5.04$\%$ and 2.68$\%$ improvements, respectively, which powerfully demonstrate that reasonable discriminate the importance of reconstructed node feature attributes from different construction paths for improving the classification performance of the existing GNN variants on graphs with features and structure missing is very important. UGCL-GCN (w/o $\mathcal{L}_{S-CL}$) denotes that we only utilize the supervision signal from the self-supervised task of feature level to guide different reconstructing paths’ model optimization. From the experimental results of the fourth and fifth columns, our proposed UGCL still achieve the best classification performance, which indicates the importance of multi-level self-supervised task for unsupervised GCL. For example, UGCL-GCN obtains a gain of 1.54 $\%$, 2.99$\%$, 0.63$\%$, 11.92$\%$, 8.96$\%$, 1.95$\%$, 3.59$\%$ and 4.69$\%$, respectively.

\begin{figure*}[ht]
\begin{minipage}[b]{0.33\linewidth}
  \centering
  \includegraphics[width=\linewidth]{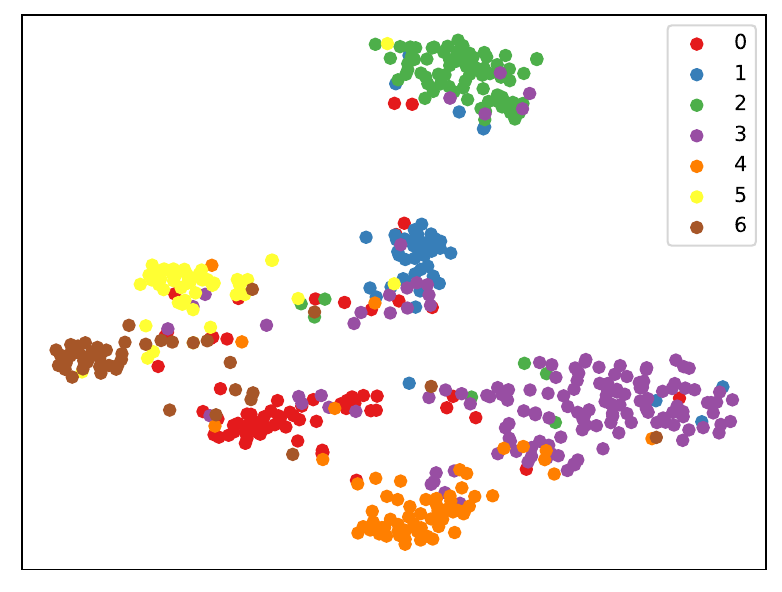}
  \centerline{(a) GCN (Cora)}\medskip
\end{minipage}
\begin{minipage}[b]{0.33\linewidth}
  \centering
  \includegraphics[width=\linewidth]{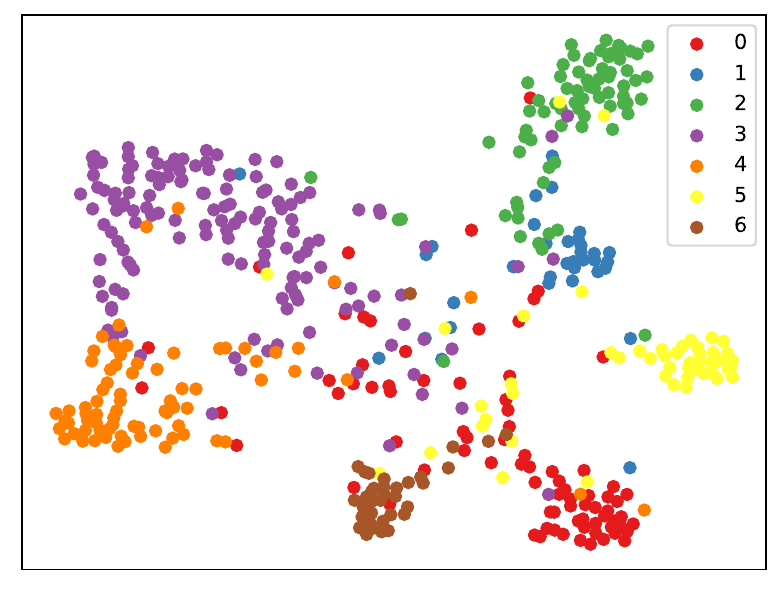}
  \centerline{(b) T2-GCN (Cora)}\medskip
\end{minipage}
\begin{minipage}[b]{0.33\linewidth}
  \centering
  \includegraphics[width=\linewidth]{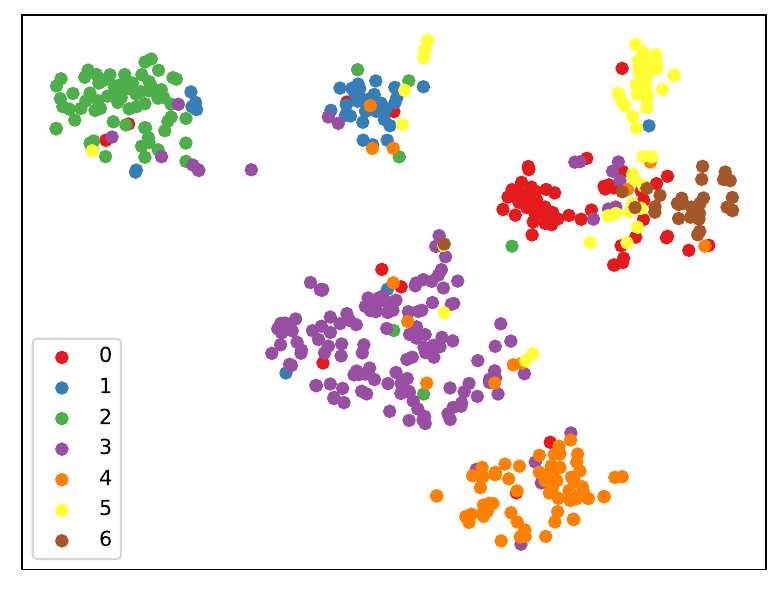}
  \centerline{(c) UGCL-GCN (Cora)}\medskip
\end{minipage}

\begin{minipage}[b]{0.33\linewidth}
  \centering
  \includegraphics[width=\linewidth]{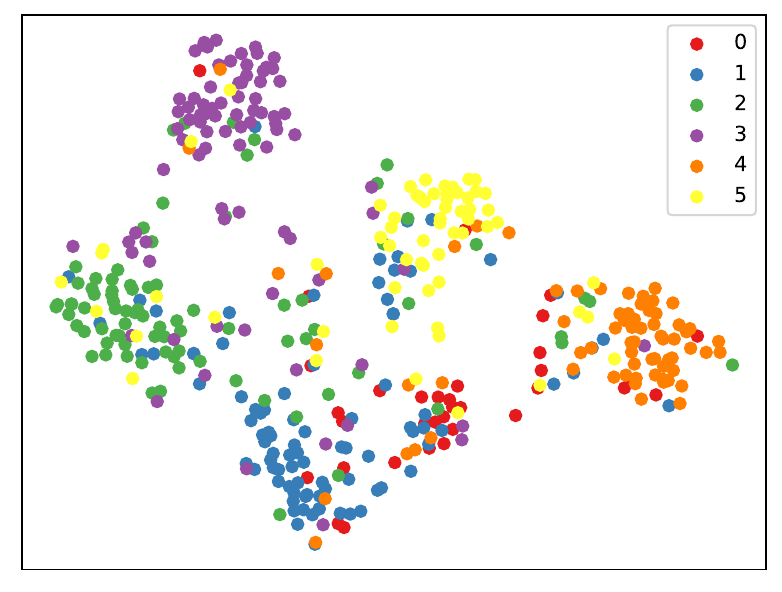}
  \centerline{(d) GCN (Citeseer)}\medskip
\end{minipage}
\begin{minipage}[b]{0.33\linewidth}
  \centering
  \includegraphics[width=\linewidth]{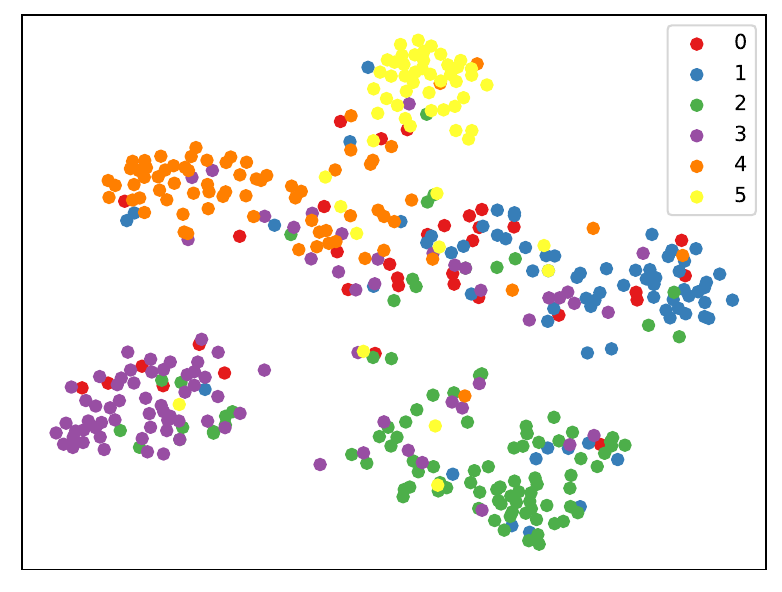}
  \centerline{(e) T2-GCN (Citeseer)}\medskip
\end{minipage}
\begin{minipage}[b]{0.33\linewidth}
  \centering
  \includegraphics[width=\linewidth]{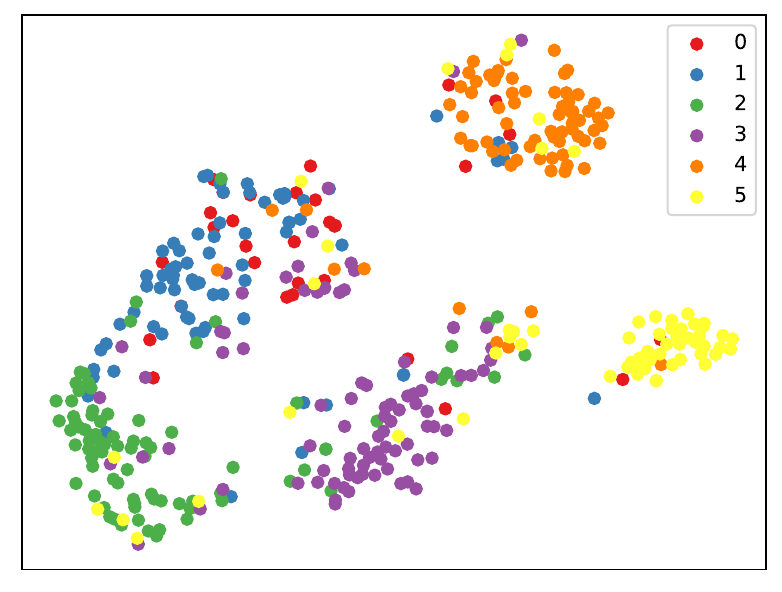}
  \centerline{(f) UGCL-GCN (Citeseer)}\medskip
\end{minipage}
\caption{t-SNE feature embeddings of different GCL methods on the Cora and Citeseer datasets with 30$\%$ missing rate. Each color represents a different class.}
\label{img}
\end{figure*}

\subsection{Effect of Different Missing Rates}

In this part, we analyze the influence of node features and structure missing rates for the classification performance of GCN, T2-GCN and UGCL methods. Table \uppercase\expandafter{\romannumeral4} reports the average accuracy of GCN, T2-GCN and UGCL with different rates of node features and structure missing. From these results, we can observe that: (1) With the node features and structure missing rates increase, the classification performance of all models goes down. Compared with GCN, the recent T2-GCN obtain superior performance in most cases and our UGCL all achieve novel state-of-the-art results. For example, in the largest Pubmed dataset, T2-GCN obtain gains of 0.14 $\%$, 0.06$\%$, 1.3$\%$, 4.45$\%$ and 1.39$\%$, UGCL achieves the 1.32 $\%$, 0.59$\%$, 2.47$\%$, 5.51$\%$ and 5.51$\%$ improvements. These results effectively demonstrate that the reconstructed node features and structure by GCL can alleviate the model collapse effect of GNN variants on graphs with features and structure missing. (2) The performance gap of T2-GCN and GCN will reduce when the downstream node classification task has a higher missing rate, such as in the 75$\%$ and 95$\%$ missing rates of the Wisconsin, Chameleon, and Squirrel datasets. In some cases, even the classification performance of T2-GCN is lower than GCN. However, our proposed UGCL still obtains a large improvement regardless of whether the node features and structure missing rates are high or low, which direct validates the generalization and effectiveness of our proposed UGCL framework.

\subsection{Effect of Different GNN variants}

To further show the scalability of the proposed UGCL on graphs with features and structure missing, we replace the adopted GCN model in the downstream node classification with other GNN variants (GraphSAGE and GAT), and then report their corresponding average accuracy on semi-supervised node classification with 30 $\%$ missing rate. Besides, we also show the classification performance of recent state-of-the-art T2-GNN with the different GNN variants. As the results show, our UGCL still obtains the best classification performance on 5 out of 8 benchmarks, which reveals that the effectiveness of our proposed method is not affected by the adopted GNN variants in the downstream node classification. Under the same experimental setting, our UGCL-GAT obtains a better performance in most cases in comparison to UGCL-GCN and UGCL-GAT, for example, UGCL-GAT outperforms UGCL-GCN by 0.61 $\%$, 0.07$\%$, 3.94$\%$, 6.21$\%$, 13.17$\%$, 1.98$\%$ and 2.48$\%$ improvements. Thus, it is also important to select its optimal GNN variants according to the different benchmarks.

% \subsection{Parameters Sensitivity}

\subsection{t-SNE Visualization}

To better demonstrate the superiority of the proposed UGCL in improving the generalization ability of the existing GNN variants on graphs with features and structure missing, we show the t-SNE feature embeddings of GCN, recent T2-GCN and our UGCL models on the Cora and Citeseer datasets with 30$\%$ missing rate. For t-SNE, sample features belonging to the same class are expected to cluster together. From Figure 4, we can see that the proposed UGCL obtains the best clustering performance and also make the separation between intra-class and inter-class clearer. Compared with GCN, the reconstructed node features and structure by UGCL let GCN learn more effective parameters. These results also indicate that our method can effectively improve the accuracy of the reconstructed node features and structure in comparison to supervised T2-GCN. 

\section{Conclusion}

Graph completion learning (GCL) has recently received widespread attention and also have been demonstrated as a successful technique in dealing with a large body of graph analytical tasks with features or structure missing. However, the superior performance of these developed GCL models highly relies on the label numbers of a specifically supervised task. Besides, their performance will have a huge drop when generalizing them to tackling the graph analytical tasks with features and structure missing, not simple features or structure missing. In this paper, we propose a more general GCL (UGCL) framework, which can effectively solve the labels reliance and the bias of the reconstructed node features and structure relationships problems that existed in the supervised GCL under the guide of self-supervised learning. In specific, considering the misleading inference between missing node features and structure caused by the message-passing scheme of GNN, we separate the feature reconstructing and structure reconstructing. Second, we complete the missing information of the original data based on the above two different reconstructing paths, respectively. A dual contrastive loss on the structure level and feature level is furtherly introduced to guide their reconstruction process by maximizing the consistency of node representations from feature reconstructing and structure reconstructing paths. Finally, the reconstructed node features and structure can be applied to any existing GNN variants for their optimization. To demonstrate the effectiveness of the proposed method, we conduct extensive experiments on the eight datasets, three GNN variants and five missing rates. Extensive results show that our proposed UGCL achieves competitive results in comparison to the existing GCL and GNN variants.

% \section*{Acknowledgment}
% This work was supported in part by the National Key Research and Development Program of China under Grant 2022YFF0712300, in part by the National Natural Science Foundation of China under Grant 62172177, in part by the Fundamental Research Funds for the Central Universities (HUST) under Grant 2022JYCXJJ034, in part by the Open Research Fund from Shandong Provincial Key Laboratory of Computer Network under Grant SKLCN-2021-02.

\bibliographystyle{IEEEtran}
\bibliography{mybib}

\end{document}